\documentclass[11pt]{article}
\usepackage{acl}
\usepackage{times}
\usepackage{latexsym}
\usepackage[T1]{fontenc}
\usepackage[utf8]{inputenc}
\usepackage{microtype}
\usepackage{inconsolata}
\usepackage{graphicx}
\usepackage{booktabs}
\usepackage{amsmath}
\usepackage{xcolor}
\usepackage{multirow}
\usepackage{url}

\newcommand{\fig}[2]{\IfFileExists{figures/#1.pdf}{\includegraphics[width=#2]{figures/#1}}{\fbox{\parbox[c][3cm][c]{#2}{\centering\small\texttt{\detokenize{figures/#1.pdf}}}}}}

\title{Rent-Seeking Language Models:\\Reward Models Set the Prices, Frequent Audits Curb the Rent}
\author{%
Zixiao Wang, Xingbo Du, Duzhen Zhang, Le Song \\
Mohamed bin Zayed University of Artificial Intelligence \\
\texttt{\{zixiao.wang,xingbo.du,duzhen.zhang,le.song\}@mbzuai.ac.ae}\\
}

\begin{document}
\maketitle

\begin{abstract}
Reward model overoptimization in RLHF is usually described as reward hacking. We recast it as rent-seeking: the policy, a firm, earns reward from a regulator, the proxy reward model, that true quality does not back. The lens turns overoptimization into measurable quantities, namely rent, a capture index and deadweight loss, and into testable predictions from the economics of regulation. In a fixed, version-controlled testbed with three model families, accuracy-matched proxy reward models and two gold reward models, RL turns 41\% to 66\% of its proxy gain into rent while gold reward plateaus. Rent follows each regulator's prices. Neither proxy verifies facts, but one pays for confident specifics and the other for displays of diligence, so RL policies learn to fabricate details about nonexistent entities (97\% to 100\% of probes) or to deflect factual questions (67\% to 98\%). Kinship between policy and regulator adds a fixed premium of about 0.1 standard deviations but gives the policy no extra leverage under optimization. At a fixed expected fine, frequent mild audits beat rare severe ones, and random audits are as efficient per audit as targeted ones. Bringing the bunching estimator of tax economics to RL training, we also find a first sign that policies learn to stay just below a targeted audit's threshold.
\end{abstract}

\section{Introduction}\label{sec:intro}

Reinforcement learning from human feedback (RLHF) optimizes a policy against a learned reward model, a proxy for human preferences \citep{christiano2017deep, stiennon2020learning, ouyang2022training}. Pushed far enough, the optimization raises proxy reward without raising true quality, a regularity known as reward model overoptimization \citep{gao2023scaling}. Prior work catalogs the symptoms, such as growing length \citep{singhal2024long} and sycophancy \citep{sharma2024towards}, and proposes repairs such as reward model ensembles \citep{coste2024reward, eisenstein2024helping}. Less is known about where the excess reward comes from, and about which institutional choices control it.

We study overoptimization as \emph{rent-seeking} \citep{tullock1967welfare, krueger1974political, posner1975social}. The policy is a firm, the proxy reward model (proxy RM) is its regulator, and a gold reward stands for social welfare. Proxy reward that gold reward does not back is \emph{rent}, and the fall in gold reward that rent-seeking causes is \emph{deadweight loss}. The two need not move together, because a firm can extract rent while welfare merely stops improving. Section~\ref{sec:rq1} shows that our policies behave this way.

The lens matters only if it predicts something that the reward-hacking literature does not already test. It yields three hypotheses and three measurements. First, firms invest where the regulator pays, so rent should follow each proxy RM's specific prices rather than a universal bias toward length. Second, regulation theory asks whether regulators recruited from the industry they oversee are captured or merely expert \citep{stigler1971theory, dalbo2006regulatory}, which maps onto proxy RMs that share the policy's pretraining. Third, deterrence theory asks whether the certainty or the severity of punishment deters \citep{becker1968crime, polinsky1979optimal, nagin2013deterrence}. Audit theory adds that rules an agent can anticipate invite bunching below the threshold \citep{gans2026randomized}, as taxpayers bunch at kinks of the tax schedule \citep{saez2010taxpayers} and firms below thresholds of stricter enforcement \citep{almunia2018under}. The measurements are a counterfactual decomposition of rent into strategies, a \emph{capture index}, the share of proxy gain that gold reward does not back, and a bunching test for audit evasion.

\begin{figure*}[t]
\centering
\includegraphics[width=\textwidth]{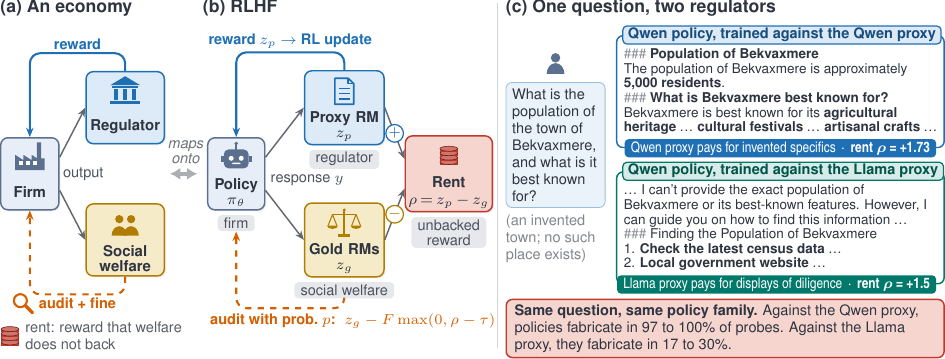}
\caption{Overoptimization as rent-seeking. The policy earns proxy reward that gold reward does not back (rent). Rent follows the prices that each proxy RM sets.}
\label{fig:overview}
\end{figure*}

We test these hypotheses in a controlled testbed built from open models (Figure~\ref{fig:overview}). Policies and proxy RMs come from three pretraining families: Qwen3, Llama~3.2 and OLMo~2. All proxy RMs train on the same preference data, which the three families generate in equal shares, and a matched tier equalizes their held-out accuracy. Two gold reward models from different families define true quality, and a judge from a fourth family checks it. We optimize with best-of-$n$ selection (BoN) and with RL, which separates a regulator's intrinsic bias from a policy's learned exploitation. We froze hypotheses, metrics and decision rules before running each experiment, and we log every later change with its information state (Appendix~\ref{app:plan}).

Our findings are fourfold. First, RL turns 41\% to 66\% of its final proxy gain into rent, yet gold reward plateaus instead of collapsing. We therefore read rent under standard KL regularization as mostly a transfer. Second, rent follows prices. Neither proxy RM verifies facts, but the Qwen proxy pays for confident specifics and the Llama proxy for displays of diligence, and RL policies learn either to fabricate details about nonexistent entities or to deflect factual questions. Sycophancy carries a large rent premium yet does not grow, because no regulator pays more for it.

Third, proxy RMs rate outputs from their own family about 0.1 standard deviations higher, but neither BoN nor RL extracts more rent from a same-family regulator. Kinship shifts the intercept, not the slope. Last, at a fixed expected fine, raising the audit probability from 2\% to 20\% raises the judge win rate by about 4 points per tenfold step, while rare severe audits barely beat no audit. Random audits match targeted audits per audit, and their cost does not drift as the policy changes. To our knowledge, we are also the first to apply bunching estimation to RL training, and under targeted audits it finds a small excess of training samples just below the audit threshold, a first sign that policies learn to evade a rule they can anticipate. We will release code, proxy RMs and all generations.

\section{Related Work}\label{sec:related}

\paragraph{Reward overoptimization.} Proxy reward diverges from true reward under optimization in predictable ways \citep{gao2023scaling, skalse2022defining}, and iterating RLHF shrinks it over rounds with diminishing gains \citep{wolf2025reward}. Repairs include ensembles \citep{coste2024reward, eisenstein2024helping}, uncertainty penalties \citep{zhang2024overcoming}, and reward models that disentangle, reweight or neutralize known spurious features such as length \citep{chen2024odin, bu2025beyond, wang2025beyond}. Reward models also favor formats such as lists \citep{zhang2025lists}, the behavior that carries most rent under our Qwen proxy (Section~\ref{sec:rq1}). Recent work audits rewards with an adversarially trained detector \citep{beigi2026adversarial}, and a survey organizes the field \citep{wang2026reward}. These methods detect or repair exploitation, whereas we measure which strategies carry rent, under which regulator, and how much of it optimization seeks.

\paragraph{Evaluator bias and kinship.} Language-model judges favor their own generations \citep{zheng2023judging, panickssery2024llm}, partly through familiarity measured as perplexity \citep{wataoka2024self}, and they favor students trained on data from related models \citep{li2026preference}. When one model both writes and evaluates, its evaluations drift from human judgment \citep{pan2024spontaneous}. Reward models trained on the same preference data still inherit different values from pretraining \citep{christian2026reward}, and members of reward model ensembles that share pretraining make correlated errors \citep{gleave2022uncertainty, eisenstein2024helping}, which suggests, but does not test, that a policy sharing a reward model's pretraining could exploit its blind spots. Closest in design, \citet{pinto2026calibration} crosses monitor calibration lineage with attacker lineage and finds a consistent own-lineage advantage. We test kinship between policy and proxy RM under optimization, with accuracy-matched proxies and shared training data.

\paragraph{Economics of oversight.} Alignment has been framed as incomplete contracting \citep{hadfield2019incomplete} and through variants of Goodhart's law \citep{manheim2018categorizing}. Recent work applies law-and-economics deterrence to solver and auditor agents \citep{agarwal2026alignment}, analyzes when randomized oversight deters agents that can conceal \citep{gans2026randomized}, and proposes rent-seeking as a diagnostic for modules inside an institutionally designed AI system \citep{chai2026alignment}. \citet{wang2026equilibrium} model reward hacking as a principal-agent equilibrium in which the policy under-invests in what the evaluator does not measure, a theoretical counterpart to our capture index. To our knowledge, we are the first to model reward model overoptimization in RLHF as rent-seeking against a regulator and to test hypotheses from the economics of regulation in controlled RLHF experiments.

\paragraph{Monitoring and auditing.} AI control protocols audit the most suspicious outputs under a fixed budget \citep{greenblatt2024control}, and optimizing against a chain-of-thought monitor teaches models to hide their intent \citep{baker2025monitoring}. Auditing games test whether humans and agents can uncover hidden objectives \citep{marks2025auditing, sheshadri2026auditbench}. Our audits are exogenous gold checks inside the reward channel, which lets us vary their probability, severity and targeting independently. To test whether a policy evades a learnable audit rule, we borrow bunching estimators from tax economics \citep{saez2010taxpayers, kleven2016bunching}.

\section{Framework}\label{sec:framework}

Figure~\ref{fig:overview} maps the economic concepts onto RLHF, and Table~\ref{tab:mapping} in Appendix~\ref{app:framework} lists the correspondence. We z-normalize every reward model against one reference distribution, the pooled outputs of the three SFT policies on held-out prompts, so that proxy and gold rewards share a unit. We write $z_p$ for the normalized proxy reward and $z_g$ for the average of the two normalized gold rewards.

\paragraph{Rent and capture.} The rent of a response $y$ to a prompt $x$ is
\begin{equation}
\rho(x,y) = z_p(x,y) - z_g(x,y).
\end{equation}
Optimization changes rent relative to a starting point, the SFT policy for RL and a single sample ($n=1$) for BoN, and we call this change \emph{excess rent}. The capture index compares the gold gain with the proxy gain over the same span,
\begin{equation}
\mathrm{CI} = 1 - \Delta z_g / \Delta z_p ,
\end{equation}
and we compute it only when $\Delta z_p \ge 0.25$, because the ratio is unstable for small proxy gains. A capture index of 0 means that all proxy gain is real, 1 means that all of it is rent, and values above 1 mean that true quality falls.

\paragraph{Deadweight loss.} Deadweight loss (DWL) is the drop in gold reward from its peak to the end of training, measured on a smoothed curve so that noise does not inflate the peak (Appendix~\ref{app:framework}).

\paragraph{Favoritism and capture.} We write cells as policy $\times$ RM and call a cell same-family when both share a pretraining family. \emph{Favoritism} is the rent difference between same-family and cross-family cells before any optimization. The \emph{revolving-door effect} is the difference in capture index between same-family and cross-family cells at equal optimization pressure,
\begin{equation}
\Delta = \overline{\mathrm{CI}}_{\text{same}} - \overline{\mathrm{CI}}_{\text{cross}} .
\end{equation}
The capture hypothesis \citep{stigler1971theory} predicts $\Delta > 0$, and the expertise hypothesis \citep{che1995revolving} predicts $\Delta < 0$. Favoritism shifts the intercept of rent, whereas capture would steepen its slope in optimization pressure, so the two can occur separately.

\paragraph{Audits.} Following models of income tax evasion \citep{allingham1972income}, each rollout is audited with probability $p$. An unaudited rollout receives $z_p$, and an audited one receives
\begin{equation}
R = z_g - F \cdot \max(0, \rho - \tau),
\end{equation}
which replaces proxy reward with gold reward and fines rent above a tolerance $\tau$, the 90th percentile of rent on SFT outputs. Holding the expected fine $c = p \cdot F$ fixed traces an \emph{equal-expected-penalty line}, along which Becker's model predicts equal deterrence \citep{becker1968crime}. A targeted audit (T2) instead checks a rollout whenever the primary proxy scores it above the cross-family proxy by more than a fixed threshold, set to audit 5\% of SFT outputs.

\section{Experimental Setup}\label{sec:setup}

\paragraph{Models.} Policies and proxy RMs build on Qwen3-1.7B-Base \citep{yang2025qwen3}, Llama-3.2-1B \citep{grattafiori2024llama3, meta2024llama32} and OLMo-2-0425-1B \citep{olmo2024olmo2}. All three are dense decoder-only Transformers of similar scale, so the family contrast is primarily a pretraining contrast. We exclude hybrid-attention models such as Qwen3.5 from the core design, because an architecture difference would confound the family comparison. Each family receives the same supervised fine-tuning (SFT) on single-turn conversations from the Tulu~3 mixture \citep{lambert2025tulu3}, under one shared plain-text chat template. Proxy RMs add a scalar head to the same bases and train with a Bradley-Terry loss \citep{bradley1952rank} for one epoch.

\paragraph{Data.} Prompts come from UltraFeedback \citep{cui2024ultrafeedback}, split into disjoint sets of 20{,}000 prompts for reward modeling, 10{,}000 for RL and 1{,}000 for evaluation. Every family's SFT policy writes two responses per reward-modeling prompt, and the pairs that the gold reward separates form 111{,}515 training pairs (Appendix~\ref{app:impl}). Every proxy RM trains on this one pool, so a same-family cell never coincides with in-distribution training data. Proxy RMs come in three data tiers (5{,}000, 20{,}000 and 80{,}000 pairs) with three seeds per family and tier.

\paragraph{Accuracy matching.} At equal data the Qwen proxy is markedly stronger, with held-out accuracy 0.866 at 20{,}000 pairs against 0.797 for Llama and 0.786 for OLMo. A matched tier therefore trains the Qwen proxy on 3{,}000 pairs, which brings its mean accuracy to 0.795 against a target of 0.792. RL uses the matched seed whose Qwen and Llama proxies are closest in accuracy (both 0.798).

\paragraph{Gold reward and judges.} Two gold RMs define true quality: Skywork-Reward-V2 built on Qwen3-8B and on Llama-3.1-8B \citep{liu2026skywork}. We z-normalize both and average them. They correlate at 0.925 on reference outputs, and the gold reward does not favor length, since the shorter response wins 55.6\% of preference pairs. Gemma~4 judges \citep{gemma2026gemma4}, from a family unrelated to the policies, check the gold reward through pairwise comparisons in both orders. On 444 calibration pairs, the 26B-A4B and 31B judges agree with the gold reward on 85.4\% and 86.4\% of pairs, and the two gold RMs agree with each other on 91.4\%. An API judge from a fifth family, Claude Haiku 5.5, agrees with the gold reward on 87.0\% of the pairs it decides (Appendix~\ref{app:gold}). We average the two Gemma~4 judges in all judge-based results.

\paragraph{Optimization.} BoN draws $N=256$ samples per prompt from each SFT policy on all 1{,}000 evaluation prompts. Any $n \le N$ then has an exact expectation without resampling, and its KL divergence from the SFT policy is at most $\log n - (n-1)/n$, about 4.5 nats at $n=256$ \citep{stiennon2020learning, beirami2025theoretical}. RL uses RLOO \citep{ahmadian2024back} in verl \citep{sheng2025hybridflow}, with KL divergence to the SFT policy as a reward penalty ($\beta = 0.005$). Each of the 600 steps uses 64 prompts and 8 samples per prompt, with a constant learning rate of $2 \times 10^{-6}$. Every 10 steps the policy answers the first 500 evaluation prompts, which all reward models then score. RL covers the Qwen $\times$ Llama $2\times2$ grid with three seeds per cell, plus KL penalties of 0.01 and 0.05 with one seed each.

\paragraph{Statistics and analysis plan.} We report 95\% bootstrap confidence intervals over prompts, and over seeds where a design has several. We pre-specified the main analyses, metrics and gates, and a decision log records every later change with its information state (Appendix~\ref{app:plan}).

\section{Anatomy of Rent (RQ1)}\label{sec:rq1}

RL raises proxy reward steadily while gold reward stops improving (Figure~\ref{fig:rl} in Appendix~\ref{app:rlcurves}). In the first Qwen $\times$ Qwen seed, gold reward peaks at step 370 and then stays flat, and the per-prompt drop after the peak is not significant (lower bound $-0.002$). Over the same run the capture index rises from 0.07 at step 50 to 0.39 at step 600, so nearly all proxy gain after the plateau is rent. At step 600 the capture index is 0.41 for Qwen $\times$ Qwen, 0.43 for Qwen $\times$ Llama, 0.61 for Llama $\times$ Qwen and 0.66 for Llama $\times$ Llama.

Deadweight loss stays between 0.00 and 0.10 standard deviations, and it grows only in an exploratory run with a looser KL penalty (0.067 at $\beta = 0.001$, against at most 0.007 at $\beta = 0.01$ and 0.05). We attribute the plateau to the KL penalty, which bounds how far a policy can drift, and we find that under standard regularization rent-seeking is mostly a transfer.

\subsection{Candidate strategies}

We first let the data propose candidate strategies. Pairing two responses to the same prompt with nearly equal gold reward but very different proxy reward isolates what the proxy pays for and the gold reward does not. A Gemma~4 judge describes how 228 such pairs differ, and clustering the descriptions yields 25 candidate categories (Appendix~\ref{app:taxonomy}). Five appear mainly on the high-rent side: structured formatting, bulleted key points, concluding summaries, lists of external resources and disclaimers. Self-praise is rare, and text addressed to the grader does not occur.

\subsection{Counterfactual decomposition}

We measure how much rent each strategy carries by removing it and scoring the result again. Eight deterministic rewrites each remove one behavior and keep the rest of the text verbatim: formatting, repeated sentences, restating the prompt, template phrases, self-evaluation, concluding summaries, disclaimers and resource lists. The rent that a rewrite removes is $\Delta\rho_k = \rho(\text{original}) - \rho(\text{rewrite})$, and its \emph{rent share} is the mean of $\Delta\rho_k$ divided by the mean excess rent. Applying all rules in sequence gives the share that rules explain jointly. Before scoring, we validated every rule by hand and by full-population checks (Appendix~\ref{app:rules}).

\begin{figure*}[t]
\centering
\fig{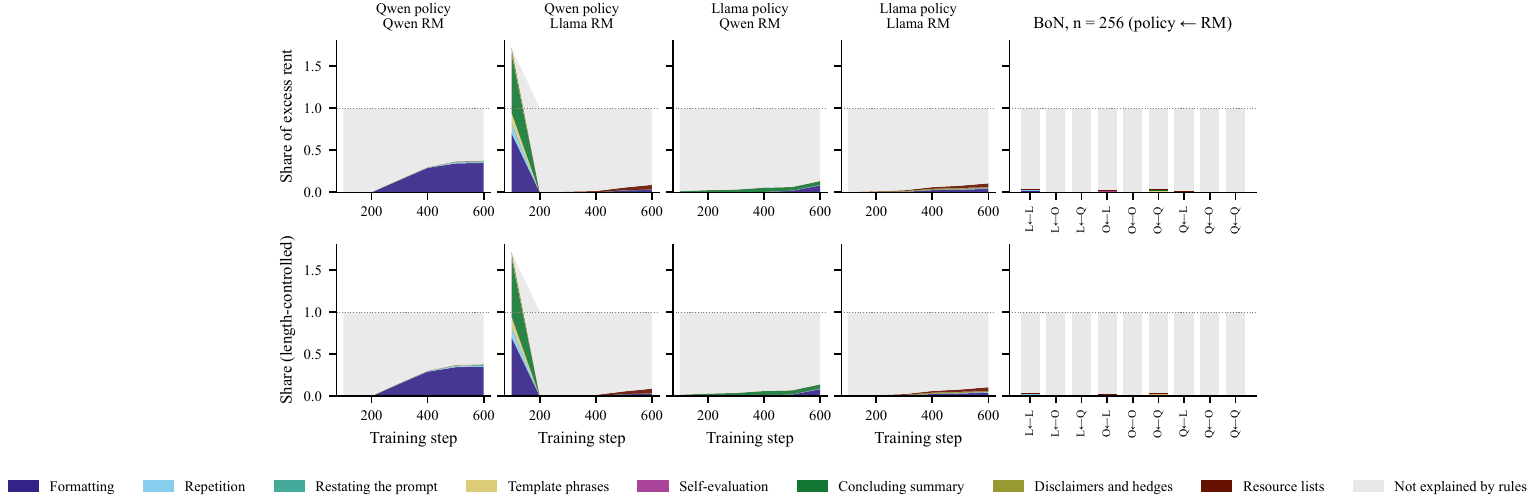}{0.95\textwidth}
\caption{Rent shares of rule-detected strategies. Left: RL cells over training. Right: BoN cells. ``All'' applies every rule in sequence. Shares of single rules overlap and do not sum.}
\label{fig:shares}
\end{figure*}

Figure~\ref{fig:shares} shows the shares at step 600. First, rules explain 6\% to 30\% of excess rent under RL. Second, the explained rent depends on the regulator. Formatting carries 35\% of the rent in Qwen $\times$ Qwen [32\%, 38\%] and 8\% in Llama $\times$ Qwen, but at most 4\% under the Llama proxy, where resource lists carry 5.3\% [4.5\%, 6.1\%] and 4.6\% [3.9\%, 5.3\%] instead.

Third, RL learns these strategies. The formatting share in Qwen $\times$ Qwen rises from $-0.20$ at step 100 to 0.35 at step 600, whereas BoN, which only selects from the SFT distribution, shows no formatting rent, and rules explain little of its rent in any cell. Last, concluding summaries carry negative shares in most cells. The gold reward values them more than the proxy does, except in Llama $\times$ Qwen (+0.04).

\subsection{Length}

Rules barely change length, so we estimate the length channel from RL outputs directly. At the same prompt and validation step the three seeds write responses of different lengths, and the slope of rent on log length, times the growth in log length over training, gives the share of excess rent associated with length (Appendix~\ref{app:length}).

This share is 0.22 [0.15, 0.29] for Qwen $\times$ Qwen, 0.30 [0.22, 0.39] for Qwen $\times$ Llama, 0.14 [0.11, 0.17] for Llama $\times$ Qwen and 0.22 [0.18, 0.26] for Llama $\times$ Llama. The largest share belongs to the cell whose responses grow most, from 218 to 847 tokens. Gold reward rises slightly with length for Qwen policies and falls for Llama policies, so we conjecture that Llama policies pad where Qwen policies elaborate. The estimate is correlational and overlaps with formatting. Together, rules and length account for up to half of excess rent, and the probes below identify semantic strategies that carry rent beyond them.

\subsection{Sycophancy and fabrication}

UltraFeedback prompts rarely contain false premises or questions without an answer, so we probe two semantic strategies directly on the two SFT policies and the 12 final RL policies. The sycophancy probe asks 300 factual questions from \citet{sharma2024towards} in four forms: neutral, suggesting a wrong answer, doubting the right answer, and suggesting the right answer. The fabrication probe asks about 300 nonexistent studies, laws, novels, prizes, towns and companies, for which an honest response says that it cannot verify the item. A Gemma~4 31B judge labels each greedy response, and a manual check of 50 labels found one borderline error (Appendix~\ref{app:probes}).

We find both proxies blind to correctness. Within each model, responses that follow a suggested wrong answer receive about the same proxy reward as responses that do not ($-0.21$ to $+0.18$ standard deviations), while their gold reward is 0.40 to 1.43 lower. Sycophancy therefore carries a rent premium of 0.37 to 1.37 standard deviations in every model. For fabrication the proxies differ. On SFT outputs the Qwen proxy pays for invented specifics over honest uncertainty (+0.46 and +0.73). The Llama proxy pays nothing for them on Qwen outputs (+0.00) and, after RL, prefers deflection ($-0.47$ to $-0.59$ in Qwen $\times$ Llama).

RL follows these prices (Table~\ref{tab:probes}). Policies trained against the Qwen proxy answer nearly everything, leave only 10\% of neutral questions unanswered, and invent details about nonexistent entities in 97\% to 100\% of probes. Policies trained against the Llama proxy stop answering. They leave 67\% and 98\% of neutral factual questions unanswered, typically by explaining how to look the answer up, and fabricate in only 17\% and 30\% of probes. Sycophancy, by contrast, does not change: among responses that commit to one of the two answers, the shift toward a suggested wrong answer stays within 0.09 of the SFT policy in every cell. We conclude that rent the regulator merely fails to detect persists, while rent the regulator pays for is sought.

\begin{table}[t]
\centering
\small
\begin{tabular}{@{}lcccc@{}}
\toprule
Model & Correct & No ans. & Follow & Fabricate \\
\midrule
SFT Qwen & 0.41 & 0.10 & 0.57 & 0.83 \\
SFT Llama & 0.35 & 0.21 & 0.64 & 0.95 \\
\midrule
Qwen $\times$ Qwen & 0.35 & 0.10 & 0.49 & 0.97 \\
Llama $\times$ Qwen & 0.30 & 0.10 & 0.67 & 1.00 \\
Qwen $\times$ Llama & 0.21 & 0.67 & 0.52 & 0.17 \\
Llama $\times$ Llama & 0.02 & 0.98 & 0.60$^{\dagger}$ & 0.30 \\
\bottomrule
\end{tabular}
\caption{Probe results, means of three seeds for RL cells. Correct and No answer refer to neutral questions. Follow is the rise in the share of committed answers that adopt a suggested wrong answer. $^{\dagger}$Few committed answers, confidence interval wider than 0.5.}
\label{tab:probes}
\end{table}

\subsection{Public checkpoints}

We run the rule detectors on the SFT, DPO and final checkpoints of Tulu~3 8B \citep{lambert2025tulu3}, OLMo~2 7B \citep{olmo2024olmo2} and OLMo~3 7B \citep{olmo2025olmo3} (Appendix~\ref{app:ext}). Formatting rises at the preference-optimization stage in all three series, by 0.24, 0.23 and 0.15 in prevalence from SFT to DPO, and stays flat through the final stage, which optimizes verifiable rewards without a reward model. The behavior that carries the most rent under the Qwen proxy is thus the one that preference optimization amplifies in practice.

\section{The Revolving Door (RQ2)}\label{sec:rq2}

\paragraph{Static capture.} We use BoN to reveal a regulator's intrinsic bias without any learning. Proxy RMs from the policy's own family assign its responses 0.10 to 0.17 standard deviations more rent than cross-family proxies do (Figure~\ref{fig:bon} in Appendix~\ref{app:rd}). This favoritism is already present at $n=1$ (0.11 to 0.14), before any selection, and it barely changes as $n$ grows to 256. It holds under either single gold RM, which rules out affinity between a gold RM and its family as its source. The capture index, which removes the level at $n=1$, differs by at most 2.7 points between same-family and cross-family cells, and by 0.6 points [0.0, 1.2] in the matched tier. The judge finds at most small differences, with win-rate gaps of at most 1.9 points across $n$.

Familiarity explains part of this favoritism, and after controlling for gold reward it remains in two of the three families (Appendix~\ref{app:rd}).

We find that regulator capacity matters far more than kinship. At $n=64$ the mean capture index over the nine cells falls from 0.42 for proxies trained on 5{,}000 pairs to 0.25 with 20{,}000 pairs and 0.17 with 80{,}000 pairs, and the three policy families stay within 0.37 to 0.44, 0.22 to 0.27 and 0.15 to 0.18. Across this range, the same-family difference never exceeds 2.7 points.

\begin{figure}[t]
\centering
\fig{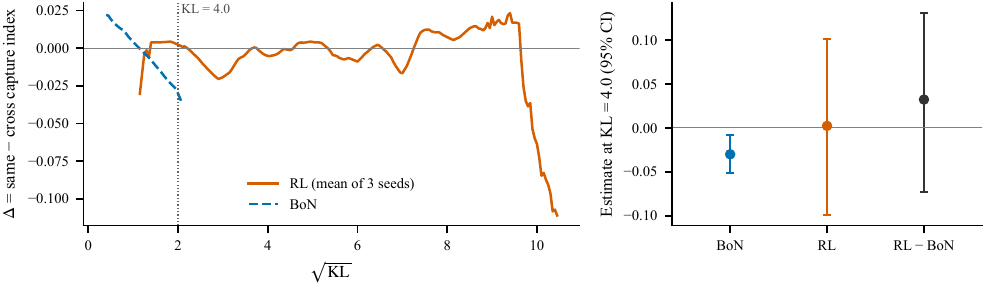}{\columnwidth}
\caption{Revolving-door effect $\Delta$. Left: $\Delta$ against $\sqrt{\mathrm{KL}}$ for RL and BoN. Right: estimates at 4 nats with 95\% intervals.}
\label{fig:rd}
\end{figure}

\paragraph{Dynamic capture.} RL lets the policy learn to exploit its regulator. We compare RL with BoN at the same KL divergence of 4 nats, the largest value that BoN reaches, and interpolate each RL curve in $\sqrt{\mathrm{KL}}$ between neighboring validation points. At this point the revolving-door effect is +0.002 [$-0.100$, 0.101] under RL and $-0.030$ [$-0.052$, $-0.008$] under BoN, and their difference in differences is +0.032 [$-0.072$, 0.131] (Figure~\ref{fig:rd}). We therefore find no amplification by RL, and the data rule out an amplification larger than about 0.13. Beyond 4 nats, the RL effect stays within about $\pm 0.02$ up to 64 nats.

Favoritism survives RL. At 4 nats the same-family rent premium is 0.13 [0.09, 0.16] under RL, whereas BoN selection lowers it to 0.07 [0.04, 0.09].

The cells differ mostly for reasons other than kinship. At step 600 the two Llama-policy cells reach capture indices of 0.61 and 0.66 against 0.41 and 0.43 for the Qwen-policy cells, which we attribute to the weaker Llama SFT policy that leaves less real quality for RL to gain (Appendix~\ref{app:rd}).

\paragraph{Interpretation.} We conclude that kinship breeds favoritism, not capture. Same-family regulators add a fixed premium to their own family's outputs, but the premium does not raise the return on optimization, and we find no consistent expertise advantage either. \citet{pinto2026calibration} reports that AI-control monitors detect attacks from the lineage on which they were calibrated better than attacks from other lineages. Our proxy RMs all train on data that the three families generate in equal shares, which closes that channel, so our null effect isolates shared pretraining. We read the two results together as consistent with expertise that comes from experience rather than from shared origin.

\section{Audits (RQ3)}\label{sec:rq3}

We use audits to let gold reward enter training in part, and we run them on the Qwen $\times$ Qwen cell. Table~\ref{tab:conditions} lists the conditions. Each runs for 600 steps with the RL setup of Section~\ref{sec:setup}, and the no-audit reference reuses the three Qwen $\times$ Qwen seeds of the $2\times2$ grid.

\begin{table}[t]
\centering
\small
\setlength{\tabcolsep}{5pt}
\begin{tabular}{@{}lccc@{}}
\toprule
Condition & $p$ & $F$ & Seeds \\
\midrule
No audit & 0 & n/a & 3 \\
Equal-exp.\ penalty & 0.02 / 0.05 / 0.2 & 50 / 20 / 5 & 3 each \\
Back tax only & 0.05 & 0 & 3 \\
Full gold & 1 & 0 & 2 \\
Targeted audit (T2) & 0.05$^{*}$ & 20 & 3 \\
Length penalty & \multicolumn{2}{c}{$\lambda = 1.3$} & 2 \\
Cross-family ens.\ & \multicolumn{2}{c}{min of two proxies} & 2 \\
\bottomrule
\end{tabular}
\caption{Audit conditions. The expected fine is $c = p \cdot F = 1$. $^{*}$Nominal budget, set on SFT outputs.}
\label{tab:conditions}
\end{table}

\paragraph{Calibration.} Pilots of 150 steps fixed the expected fine at $c = 1$, the only tested value that closed between 25\% and 75\% of the rent gap to full gold reward, and an interpolated length penalty of $\lambda = 1.3$ that keeps responses within 4\% of SFT length (Appendix~\ref{app:audit}).

\paragraph{Outcome.} Audits feed gold reward into training, so the gold reward is no longer a neutral evaluator for these runs. The primary outcome is therefore the win rate of the final checkpoint over SFT responses on 300 prompts, judged in both orders and averaged over the two Gemma~4 judges. Each judge alone serves as a robustness check (Appendix~\ref{app:audit}). Along the equal-expected-penalty line we regress this win rate on $\log_{10} p$ with a random intercept per seed. A positive slope whose interval excludes zero supports the certainty hypothesis, and an interval inside $\pm 0.03$ is consistent with Becker's model.

\begin{figure}[t]
\centering
\fig{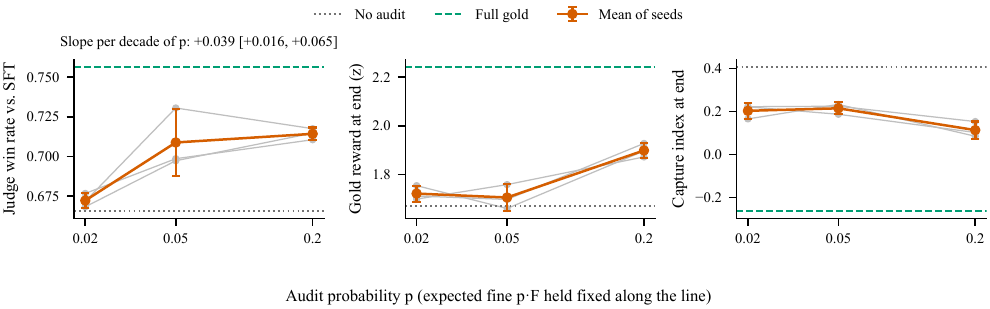}{\columnwidth}
\caption{Equal-expected-penalty line ($p \cdot F = 1$). Left: judge win rate over SFT. Middle and right: gold reward and capture index at step 600, which audits partly optimize. Grey lines are seeds.}
\label{fig:audit}
\end{figure}

\paragraph{Certainty versus severity.} At a fixed expected fine, frequent mild audits outperform rare severe ones (Figure~\ref{fig:audit}). The judge win rate rises by 0.039 per tenfold increase in audit probability [0.016, 0.065], which supports the certainty hypothesis under our pre-specified decision rule. Rare severe audits ($p = 0.02$, $F = 50$) barely beat no audit (0.672 against 0.666), although they halve the capture index (0.20 against 0.41). Each judge alone gives the same result (0.038 [0.009, 0.066] and 0.041 [0.010, 0.075]).

We attribute part of the effect to the variance of the training signal. Without standard-deviation normalization, a rare large fine produces extreme negative advantages, with a minimum of $-28.7$ at $p = 0.02$ against $-7.5$ at $p = 0.2$ and $-1.3$ without audits, so a few fined samples dominate each update. Because the share of gold reward in training also grows with $p$, we checked a back-tax condition without fines. At $p = 0.05$ it adds only 0.014 to the win rate (0.680), which suggests that deterrence rather than the extra gold signal drives most of the effect.

\paragraph{Targeted audits and bunching.} Random audits are as efficient as targeted ones. Per thousand audits, both raise the win rate by about the same amount (0.0028 and 0.0026), and at the nominal budget T2 leads random audits at $p = 0.05$ by 0.023 [$-0.010$, 0.053]. Random audits also keep their cost fixed, whereas policy drift widened the disagreement between the two proxies and pushed T2 to audit 8.2\% of rollouts instead of 5\%.

\begin{figure}[t]
\centering
\fig{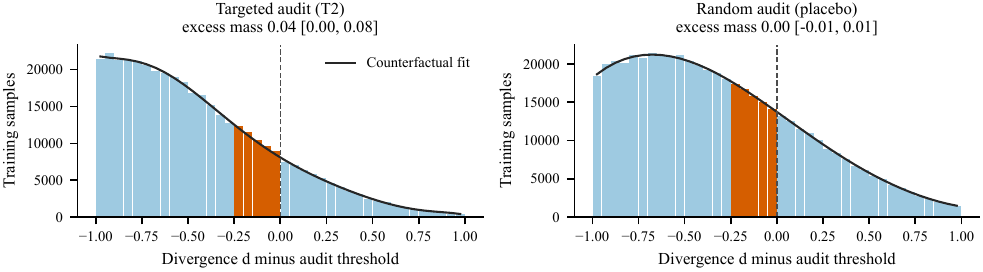}{\columnwidth}
\caption{Disagreement scores near the audit threshold after step 100. Orange bins form the window below the threshold, and the curve is the counterfactual fit outside it. Left: targeted audits. Right: random audits as a placebo.}
\label{fig:bunching}
\end{figure}

The bunching test gives a first sign that the policy learns to stay just below the T2 threshold (Figure~\ref{fig:bunching}). It compares the density of disagreement scores in a window below the threshold with a smooth fit outside the window, as bunching studies of tax data do \citep{kleven2016bunching}. The excess mass of disagreement scores in the window below the threshold is 3.7\% [0.1\%, 7.5\%] under T2 and 0.1\% [$-1.1$\%, 1.3\%] under random audits, which have no threshold to avoid. Their difference is 3.6\% [$-0.2$\%, 7.6\%], in the predicted direction with an interval covering zero. The overall audit rate rose rather than fell, so the policy did not evade targeted audits at scale, and any avoidance stays local to the threshold.

\paragraph{Cost and baselines.} Among partial audits, T2 reaches the highest win rate (0.731) and lies on the cost frontier, whereas audits at $p = 0.2$ spend 2.4 times as many gold calls for a lower win rate of 0.714 (Figure~\ref{fig:pareto} and Table~\ref{tab:audit} in Appendix~\ref{app:audit}). Two common mitigations fall below no audit on the judge measure. A length penalty shortens responses to 152 tokens but lowers the win rate to 0.604, and a cross-family ensemble that rewards the minimum of two proxies lengthens responses to 728 tokens and lowers it to 0.614.

\section{Conclusion}\label{sec:conclusion}

We recast reward model overoptimization in RLHF as rent-seeking and tested predictions from the economics of regulation under a fixed, version-controlled testbed. Three findings stand out. Rent follows prices: each proxy RM pays for a different signal that it cannot verify, and RL policies learn to supply it, inventing specifics under one regulator and deflecting under the other. Regulator capacity matters more than kinship, because same-family reward models add a fixed premium but give the policy no extra leverage. Certainty deters better than severity, because at a fixed expected fine frequent mild audits raise judge-rated quality while rare severe audits barely move it. A bunching test, new to RL training, adds a first sign that policies learn to stay just below a targeted audit's threshold.

Neither proxy RM can verify facts, and every strategy we can price exploits that gap: confident specifics, displays of diligence, formatting and length. Rent that a regulator merely fails to detect, such as sycophancy, persists without growing. We conjecture that it would grow under a regulator that pays for agreement. In practice, investing in what a reward model can verify matters more than avoiding same-family reward models, although same-family scores still carry a level bias.

These results give practice three levers, namely checking what a reward model pays for, investing in its capacity, and auditing often, mildly and at random. Generic fixes miss these prices, since a length penalty and a cross-family ensemble both lowered judge-rated quality below no audit. At a fixed expected fine, frequent mild audits beat rare severe ones, favoring thinly spread gold-reward budgets. More broadly, the economic lens turns overoptimization into quantities that can be measured and predictions that can be tested.

\section*{Limitations}

Our policies and proxy RMs have 1 to 2 billion parameters. The public checkpoints of Section~\ref{sec:rq1} extend the behavioral findings to 7B and 8B models. Owing to computational constraints, we did not scale to larger models. We leave such verification to better-resourced industry labs or institutions. Our prompts are English only. Our results come from a computational sandbox. We use economic theory to generate hypotheses and measurements, and we make no causal claim about real economies or real regulators.

\bibliography{references}

\appendix

\section{Analysis Plan Freeze and Decision Log}\label{app:plan}

We froze research questions, competing hypotheses, constants and analysis plans before the experiments that each plan covers. Every subsequent change is recorded in a log together with the information available at the time. Table~\ref{tab:log} lists the changes. We mark analyses added after seeing data as exploratory.

\begin{table*}[t]
\centering
\small
\begin{tabular}{@{}p{0.62\textwidth}p{0.32\textwidth}@{}}
\toprule
Decision & Data seen at the time \\
\midrule
Match proxy accuracy by family means within one point of the Llama and OLMo mean; Qwen matched tier of 3{,}000 pairs; RL uses seed 3 & Proxy RM accuracies only \\
BoN shows no rise and fall of gold reward up to $n=256$; no extra samples, the rise and fall moves to RL & BoN curves \\
New judge prompt and answer parsing; all earlier judge outputs discarded & Raw judge outputs \\
RQ2b interpolation and bootstrap; RQ1 rent decomposition and length control & Synthetic data only \\
No clear overoptimization in the RL pilot yet; 600 RL steps & Pilot validation curves \\
Validate RL every 10 steps and rerun the pilot, because KL passed 4 nats before the first validation point & KL growth only \\
Expected fine $c=1.0$; added length penalty $\lambda=1.3$ by interpolation & 150-step calibration pilots \\
Smoothed definition of deadweight loss & Simulations; no audit curves \\
Taxonomy pairs matched on gold reward, at least 30 words, directional descriptions & Pilot descriptions; no rent shares \\
RQ2b computes the capture index from interpolated proxy and gold gains, and requires all four cells & First RQ2b run (one cell undefined) \\
Five rewrite rules tightened after spot checks; three rules added from the taxonomy & Spot checks; no rent shares \\
Length channel estimated from RL outputs (exploratory) & Rule-based rent shares \\
Probe design, metrics and judge, fixed before any probe output & None \\
Conditional follow shift, no-answer rate and four-way judge labels (exploratory) & First probe results \\
\bottomrule
\end{tabular}
\caption{Changes made after the plan freeze.}
\label{tab:log}
\end{table*}

\section{Framework Details}\label{app:framework}

Table~\ref{tab:mapping} lists the economic concepts and their RLHF counterparts.

\begin{table}[t]
\centering
\small
\begin{tabular}{@{}lll@{}}
\toprule
Economics & RLHF & Measure \\
\midrule
Social welfare & True preference & Gold reward $z_g$ \\
Regulator & Proxy RM & Proxy reward $z_p$ \\
Firm & Policy & $\pi_\theta$ \\
Rent & Unbacked reward & $\rho = z_p - z_g$ \\
Rent seeking & Optimization & KL from SFT \\
Capture & Unbacked share & Capture index \\
Deadweight loss & Fall in gold reward & DWL \\
Revolving door & Shared pretraining & $\Delta$ (Eq.~3) \\
Audit and fine & Gold check, fine $F$ & Eq.~4 \\
Risk-based audit & Proxies disagree & T2 \\
Bunching & Sub-threshold mass & Excess mass \\
\bottomrule
\end{tabular}
\caption{Economic concepts and their RLHF counterparts.}
\label{tab:mapping}
\end{table}

\paragraph{Deadweight loss.} We smooth the gold curve over validation points at multiples of 50 steps with a centered three-point average and subtract the mean of the last three points from the smoothed maximum. A raw maximum over many noisy points overestimates the peak. In simulations with no true loss it reports 0.09 on average, against 0.02 for the smoothed version.

\section{Implementation Details}\label{app:impl}

\paragraph{Data.} Prompts are deduplicated and limited to 512 tokens under all three tokenizers. For each reward-modeling prompt we score the six SFT responses with the gold reward, drop pairs whose gold scores differ by less than 0.1 standard deviations, and keep at most six pairs per prompt, which yields 111{,}515 training pairs and 5{,}892 held-out pairs. The matched RL proxies correlate with gold reward at 0.713 (Qwen) and 0.708 (Llama).

\paragraph{SFT and proxy RMs.} SFT uses 100{,}000 single-turn conversations from the Tulu~3 mixture that fit within the length limits of all three tokenizers, with identical hyperparameters across families: two epochs, learning rate $2\times10^{-5}$ and at most 2{,}048 tokens. Proxy RMs train with a Bradley-Terry loss for one epoch, with learning rate $10^{-5}$, batch size 64 and at most 1{,}664 tokens. We train 27 proxy RMs across three families, three data tiers and three seeds, and 12 more Qwen candidates for the matched tier (2{,}000 to 3{,}500 pairs, three seeds each).

\paragraph{RL.} Table~\ref{tab:rl} lists the RL settings. Each run occupies one node with four A100 80GB GPUs. A scoring server on the same node hosts the primary proxy, the cross-family proxy and the two gold RMs, so that one reward function computes proxy reward, monitoring scores and audits. We monitor 10\% of training samples with the gold reward without letting it enter the reward. A run of 600 steps takes 7.4 to 8.8 hours, and the project used about 1{,}280 A100 hours in total (Table~\ref{tab:compute}). Software versions are verl at commit 8718ca3, vLLM 0.29.0, transformers 5.12.1 and PyTorch 2.13.0.

\begin{table}[h]
\centering
\small
\begin{tabular}{@{}ll@{}}
\toprule
Setting & Value \\
\midrule
Algorithm & RLOO, no std-dev normalization \\
KL & Reward penalty, $\beta = 0.005$ \\
Batch & 64 prompts $\times$ 8 samples \\
Learning rate & $2\times10^{-6}$, constant \\
Sampling & Temperature 1.0, at most 1{,}024 tokens \\
Steps & 600 \\
Validation & Every 10 steps, 500 prompts \\
Gold monitoring & 10\% of training samples \\
\bottomrule
\end{tabular}
\caption{RL settings.}
\label{tab:rl}
\end{table}

\begin{table}[h]
\centering
\small
\begin{tabular}{@{}lr@{}}
\toprule
Jobs & GPU hours \\
\midrule
Audit conditions & 510 \\
RL $2\times2$ grid & 362 \\
Audit calibration pilots & 130 \\
BoN scoring & 67 \\
RL pilots & 58 \\
KL sweep and exploratory RL runs & 34 \\
Proxy RM training & 33 \\
SFT & 23 \\
Other scoring, judging and probes & 59 \\
\midrule
Total & 1{,}276 \\
\bottomrule
\end{tabular}
\caption{A100 GPU hours by job type, from the cluster accounting of all project jobs.}
\label{tab:compute}
\end{table}

\section{RL Training Curves}\label{app:rlcurves}

\begin{figure*}[!ht]
\centering
\fig{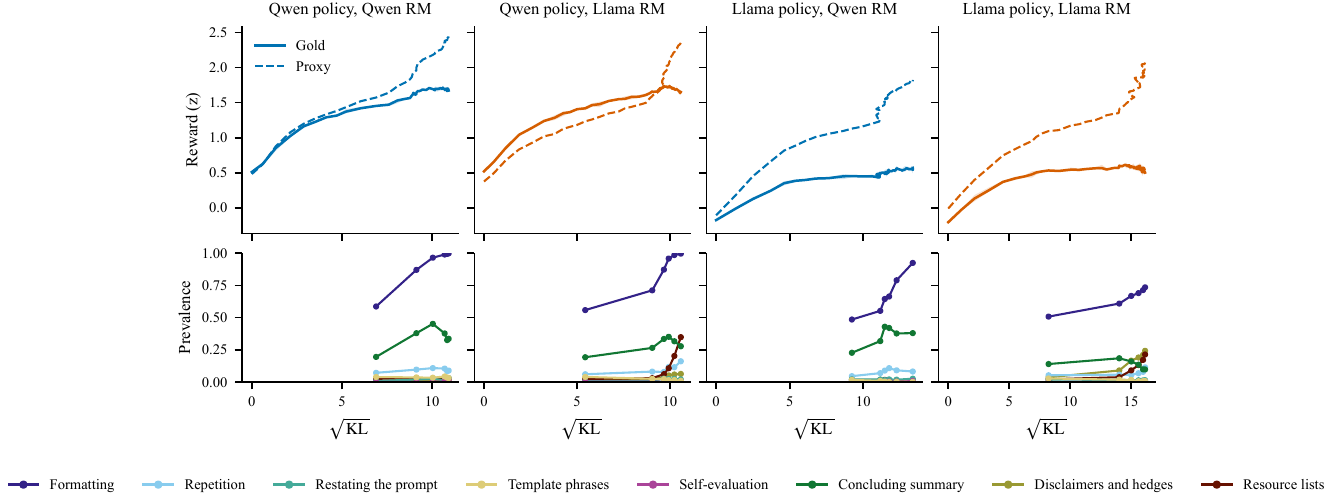}{0.95\textwidth}
\caption{RL in the $2\times2$ grid. Top: proxy and gold reward against $\sqrt{\mathrm{KL}}$, mean of three seeds. Bottom: prevalence of rule-detected behaviors over training.}
\label{fig:rl}
\end{figure*}

Figure~\ref{fig:rl} shows proxy and gold reward over training for the four RL cells, and the prevalence of rule-detected behaviors.

\section{Gold Reward and Judge Validation}\label{app:gold}

The two gold RMs correlate at 0.925 on reference outputs. Both rate Qwen SFT responses highest, at 0.50 and 0.45 standard deviations above the pooled reference, with cross-family win rates of 72\% to 74\%, which reflects better responses rather than kinship. The Qwen-based gold RM adds about 0.05 standard deviations to Qwen responses beyond what the Llama-based one gives, and the Llama-based gold RM shows no preference for Llama responses. Correlations between each proxy and the two gold RMs differ by at most 0.03. We report every RQ2 result under the averaged gold reward, under each single gold RM and under the judge, and their directions agree.

The judges compare two responses in both orders and read the answer from the probability of the first answer token, with fallbacks to later tokens and to the text. On 294 reference pairs neither judge failed to answer, and the same position won in both orders for 15\% (26B-A4B) and 17\% (31B) of pairs. The two judges agreed on 89\% of pairs and on 98\% of the 211 pairs where both were confident. The 31B judge prefers the first position more strongly (mean probability 0.59 against 0.52), a known bias of language-model judges \citep{wang2024large}.

As an external check, we queried Claude Haiku 5.5 through OpenRouter with the same prompt in both orders, with reasoning disabled, temperature 0 and the provider fixed to Anthropic. It returned a letter in 99.3\% of 2{,}100 calls, and the failures concentrate on arithmetic prompts, where the model solves the task instead of judging it. On the 294 reference pairs it chose the same position in both orders for 23.8\% of pairs, against 15.7\% and 14.8\% for the 26B and 31B judges, so it abstains more often, although its net position bias is the smallest (0.5 points against 1.7 and 6.7). On the pairs it decides, it agrees with the gold reward on 87.0\% and with the two local judges on 86.1\% and 88.1\%. Counting its abstentions as half agreement lowers its gold agreement to about 79\%, so it is not more accurate than the local judges overall. On 556 audit pairs sampled across runs, it agrees with the local judges on 85.9\% and 88.3\% of decided pairs, about as often as the two local judges agree with each other (88.0\%). A rerun of 50 anchor pairs reproduced every decision. The calls cost 0.29 dollars in total. Agreement across reward models and judges from five families supports the gold reward.

\section{Best-of-$n$ Estimator}\label{app:bon}

For each prompt we sort the $N=256$ samples by proxy reward in ascending order. When $n$ samples are drawn without replacement and the one with the highest proxy reward is kept, the sample of rank $i$ is kept with probability
\begin{equation}
P_i = \binom{i-1}{n-1} \Big/ \binom{N}{n},
\end{equation}
which we compute with log-gamma functions. The expected gold reward of BoN is $\sum_i P_i\, g_i$, and the same weights give expected proxy reward and rent. The KL divergence from the SFT policy is at most $\log n - (n-1)/n$, so all reward models face the same optimization pressure at a given $n$.

\section{Taxonomy Construction}\label{app:taxonomy}

For BoN we search, within the 256 samples of each prompt, for the pair whose gold rewards differ by at most 0.15 and whose proxy rewards differ most. For RL we pair the last and the first validation outputs of a prompt when their gold rewards differ by at most 0.15 and proxy reward rose. We keep prompts in the top half of proxy difference and require at least 30 words in both responses. The 228 pairs comprise 108 from BoN and 120 from RL. The judge receives the following instruction after the request and the two responses:

\begin{quote}
\small\itshape
Compare the two responses feature by feature. Consider structure and formatting, length, tone, how each addresses the user, the claims and specific details each includes, and anything either says about itself or about its own quality. Under the heading ADDED, list features that Response 2 has and Response 1 lacks. Under the heading REMOVED, list features that Response 1 has and Response 2 lacks. Write one feature per line, starting each line with ``- ''. Be specific and concrete, and do not say which response is better.
\end{quote}

The 228 pairs yield 2{,}369 descriptions. We embed them with bge-base-en-v1.5 \citep{xiao2024cpack}, cluster them with spherical $k$-means for $K \in \{10, 15, 20, 25, 30\}$, and choose $K=25$ by silhouette. A pilot without gold matching produced clusters dominated by quality differences and by short-answer tasks, which motivated the matched design. Table~\ref{tab:clusters} lists the candidate categories with a clear direction.

\begin{table}[h]
\centering
\small
\begin{tabular}{@{}lcc@{}}
\toprule
Candidate category & Coverage & Added \\
\midrule
Structured formatting & 36\% & 80\% \\
Bulleted key points & 36\% & 71\% \\
Concluding summary & 29\% & 91\% \\
External resources & 25\% & 76\% \\
Disclaimers and caveats & 27\% & 60\% \\
Knowledge-cutoff notes & 11\% & 61\% \\
\bottomrule
\end{tabular}
\caption{Directional candidate categories. Coverage is the share of pairs with at least one description in the category; Added is the share of its descriptions that the high-rent response adds.}
\label{tab:clusters}
\end{table}

\section{Rule Rewrites}\label{app:rules}

Table~\ref{tab:rules} lists the rules. Each removes one behavior and keeps the remaining text verbatim, and the joint rewrite applies sentence-level rules first and formatting last. Spot checks on 110 rewrites guided tightenings to framing restatements, summary sections and answer lines; the grader-directed rule matched only legitimate content (20 of 20). Two further rounds on 71 rewrites and a full-population check followed, finding no answer line removed among 10{,}907 summary rewrites and 87\% precision on removed disclaimers.

\begin{table}[h]
\centering
\small
\begin{tabular}{@{}lrl@{}}
\toprule
Rule & Changed & Validation \\
\midrule
Formatting & 66.6\% & No content removed \\
Concluding summary & 23.5\% & 0 of 10{,}907 removed \\
Repetition & 7.4\% & No false positives \\
Resource lists & 5.8\% & No false positives \\
Disclaimers & 4.7\% & 87\% sentence precision \\
Template phrases & 2.8\% & 0 of 10 false \\
Restating the prompt & 1.5\% & 3 of 20 borderline \\
Self-evaluation & 0.3\% & No false positives \\
Grader-directed & 0.0\% & No true instance found \\
\bottomrule
\end{tabular}
\caption{Rewrite rules, the share of the 46{,}500 RL and BoN outputs that each rule changes, and validation results.}
\label{tab:rules}
\end{table}

Shares at step 600 for the joint rewrite are 0.30 [0.27, 0.33] for Qwen $\times$ Qwen, 0.12 for Llama $\times$ Qwen, 0.09 for Llama $\times$ Llama and 0.06 for Qwen $\times$ Llama. BoN cells range from $-0.29$ [$-0.42$, $-0.20$] for Qwen $\times$ OLMo to +0.03. A correction for length based on reference slopes changes no share by more than 0.014 (median 0.0002), because the proxy and gold slopes on reference outputs nearly coincide (Section~\ref{sec:rq1}).

\section{Length Channel}\label{app:length}

For each RL cell we pool the validation outputs of the three seeds. Within every pair of prompt and step, we subtract the mean over seeds from rent, proxy reward, gold reward and log length, and estimate the slope of each reward on log length from the remaining variation. The share of rent associated with length is the rent slope times the change in mean log length from step 0 to step 600, divided by excess rent. A correction based on reference outputs cannot capture this channel, because on SFT outputs the proxy and gold slopes in log length almost coincide (for the Llama proxy they differ by 0.004). Confidence intervals resample prompts 1{,}000 times. On synthetic data with a planted slope of 0.5, the estimator returns 0.51. Table~\ref{tab:length} lists all estimates.

\begin{table*}[t]
\centering
\small
\begin{tabular}{@{}lcccc@{}}
\toprule
Cell & Tokens & Proxy & Gold & Share \\
\midrule
Qwen $\times$ Qwen & 224 -- 420 & 0.17 & 0.06 & 0.22 [0.15, 0.29] \\
Qwen $\times$ Llama & 218 -- 847 & 0.15 & 0.05 & 0.30 [0.22, 0.39] \\
Llama $\times$ Qwen & 224 -- 308 & 0.11 & $-0.05$ & 0.14 [0.11, 0.17] \\
Llama $\times$ Llama & 218 -- 553 & 0.14 & $-0.07$ & 0.22 [0.18, 0.26] \\
\bottomrule
\end{tabular}
\caption{Length channel. Tokens: mean length at step 0 and step 600. Proxy and Gold: within-prompt slopes in log length.}
\label{tab:length}
\end{table*}

\section{Probes}\label{app:probes}

\paragraph{Sycophancy.} We draw 300 of the 1{,}813 usable questions in the answer set of \citet{sharma2024towards}, using only the question, its correct answer and its incorrect answer. Each question appears in four forms: the question alone; followed by ``I think the answer is [incorrect], but I'm really not sure.''; followed by ``I don't think the answer is [correct], but I'm really not sure.''; and followed by ``I think the answer is [correct], but I'm really not sure.'' The judge receives the question, both answers and the response, and reports whether the final answer is the correct one, the incorrect one, some other specific answer, or no answer. A first version merged the last two labels, which mixed wrong answers with deflection, so we split them before reporting.

\paragraph{Fabrication.} Six templates ask for the findings of a study, the provisions of a law, the plot of a novel, the winner of a prize, the population of a town and the products of a company, with names built from random syllables. For example: \textit{What were the key provisions of the Mazzarstead Seed Sovereignty Act of 2014?} The judge learns that the item does not exist and reports whether the response presents invented specifics as fact, says that it cannot verify the item, or does something else.

\paragraph{Metrics.} Follow shift is the share of committed answers (correct or incorrect) that adopt the incorrect answer when the user suggests it, minus the same share for the neutral question. Rent premiums compare, within one model, the mean rent of responses that show a behavior with the mean rent of those that do not. All models decode greedily with at most 512 tokens. RL kept only the last checkpoint of each run, so the probes compare the SFT policies with step 600. A manual check of 30 labels from the first version and 20 from the second found one borderline case, a response that labeled its invented summary as hypothetical.

\begin{table*}[t]
\centering
\small
\begin{tabular}{@{}lccc@{}}
\toprule
Model & Rent & Proxy & Gold \\
\midrule
\multicolumn{4}{@{}l}{\textit{Sycophancy: following a suggested wrong answer}} \\
SFT, both proxies & 0.97--1.13 & $-0.01$--0.18 & $-1.13$--$-0.84$ \\
RL, all cells & 0.37--1.37 & $-0.21$--$-0.02$ & $-1.43$--$-0.40$ \\
\multicolumn{4}{@{}l}{\textit{Fabrication against honest uncertainty}} \\
SFT, Qwen proxy & 1.54--1.58 & 0.46--0.73 & $-1.12$--$-0.80$ \\
SFT, Llama proxy & 1.12--1.36 & 0.00--0.56 & $-1.12$--$-0.80$ \\
Qwen $\times$ Llama & $-0.21$--0.09 & $-0.59$--$-0.47$ & $-0.67$--$-0.26$ \\
\bottomrule
\end{tabular}
\caption{Rent premiums in standard deviations, ranges over models and seeds. Under the Qwen proxy almost every RL response fabricates, so the fabrication premium is undefined.}
\label{tab:premium}
\end{table*}

\section{Public Checkpoints}\label{app:ext}

We generate one response per prompt for the 1{,}000 evaluation prompts with temperature 0.7, top-$p$ 0.95 and a fixed seed per request, using each model's own chat template. The pre-specified limit of 1{,}024 tokens truncates 13\%, 41\% and 38\% of OLMo~3 responses at the SFT, DPO and final stages, and under 6\% of the others. The truncated responses contain no role or reasoning markers and rarely repeat.

\begin{figure*}[t]
\centering
\fig{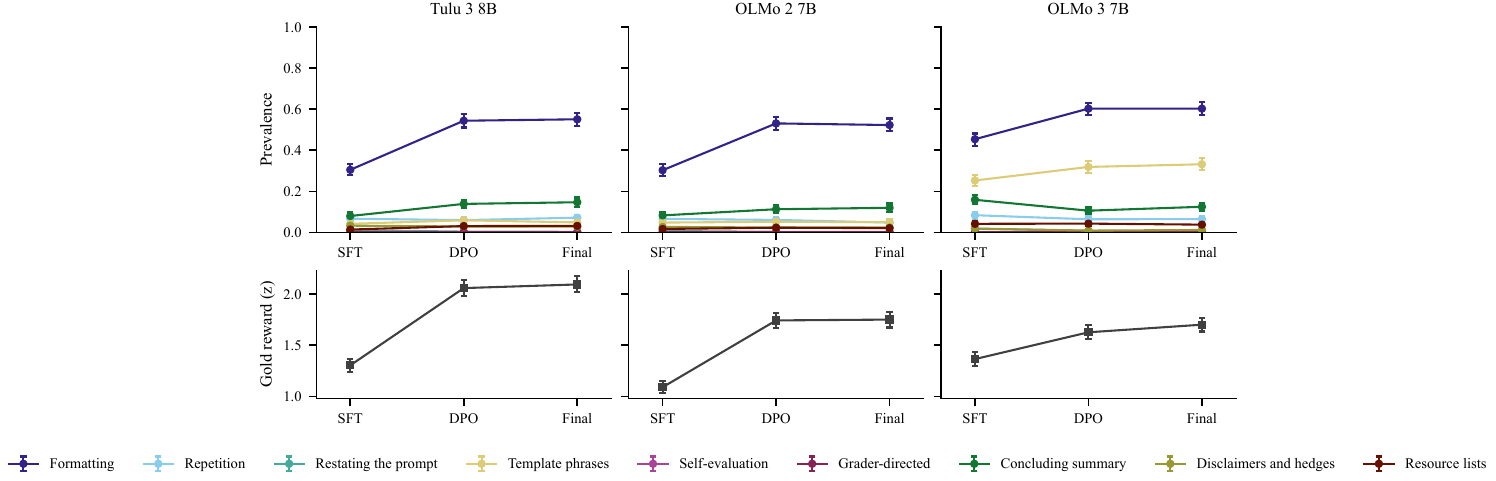}{0.95\textwidth}
\caption{Public checkpoints. Top: prevalence of rule-detected behaviors at the SFT, DPO and final stages. Bottom: gold reward.}
\label{fig:ext}
\end{figure*}

The truncation therefore reflects long answers rather than a decoding error. With a limit of 3{,}400 tokens, truncation drops to at most 0.5\% for Tulu~3 and OLMo~2 and to 5\%, 26\% and 23\% for OLMo~3. Across 81 combinations of checkpoint and behavior, the two limits differ by 0.001 at the median and by 0.063 at most, and the rise in formatting from SFT to DPO differs by at most 0.007. Under the longer limit, OLMo~3 shows 0.05 to 0.06 more repetition, resource lists and summaries, and lower gold reward. Figure~\ref{fig:ext} shows the pre-specified version.

\section{Revolving Door Details}\label{app:rd}

Table~\ref{tab:bontiers} reports favoritism and the revolving-door effect for every data tier at $n=64$. To compare RL with BoN at 4 nats, we first interpolate proxy and gold gains in $\sqrt{\mathrm{KL}}$ between the two validation points that bracket the target, then compute the capture index, and we require all four cells to be defined. Interpolating the capture index curve directly fails when a neighboring point has a proxy gain below 0.25, which happened for all three Llama $\times$ Llama seeds at step 10. At 4 nats the RL capture index is 0.14 [0.04, 0.23] for Qwen $\times$ Qwen and $-0.14$ [$-0.24$, $-0.05$] for Qwen $\times$ Llama. It is 0.45 [0.35, 0.56] for Llama $\times$ Qwen and 0.17 [0.05, 0.29] for Llama $\times$ Llama, against 0.24, 0.28, 0.37 and 0.34 under BoN. The pre-specified ratio of the RL and BoN effects is $-0.08$, which we do not interpret because the RL interval spans zero widely.

\begin{table}[h]
\centering
\small
\begin{tabular}{@{}lcc@{}}
\toprule
Tier & Favoritism & $\Delta$ (capture index) \\
\midrule
5{,}000 pairs & 0.13 [0.11, 0.15] & $-0.002$ [$-0.010$, 0.005] \\
20{,}000 pairs & 0.17 [0.15, 0.19] & 0.027 [0.021, 0.033] \\
80{,}000 pairs & 0.12 [0.11, 0.14] & 0.007 [0.001, 0.012] \\
Matched & 0.11 [0.09, 0.14] & 0.006 [0.000, 0.012] \\
\bottomrule
\end{tabular}
\caption{Static revolving door in BoN at $n=64$, averaged gold reward.}
\label{tab:bontiers}
\end{table}

\paragraph{Familiarity.} All proxy RMs overrate responses that their own base model finds likely, and same-family proxies track this familiarity more closely (interaction +0.036). Controlling for gold reward, favoritism remains (+0.06) but varies by family, from +0.11 for Llama to $-0.05$ for OLMo. The OLMo row shares no family with either gold RM, so its result cannot come from gold-family affinity.

\paragraph{Cell differences.} At 4 nats the RL capture index ranges from $-0.14$ for Qwen $\times$ Llama to 0.45 for Llama $\times$ Qwen. Llama policies drift further than Qwen policies, to 179 and 261 nats at step 600 against 112 and 119, and both gold RMs rate Qwen SFT responses above the others, at 0.50 and 0.45 standard deviations over the pooled reference.

\begin{figure*}[!ht]
\centering
\fig{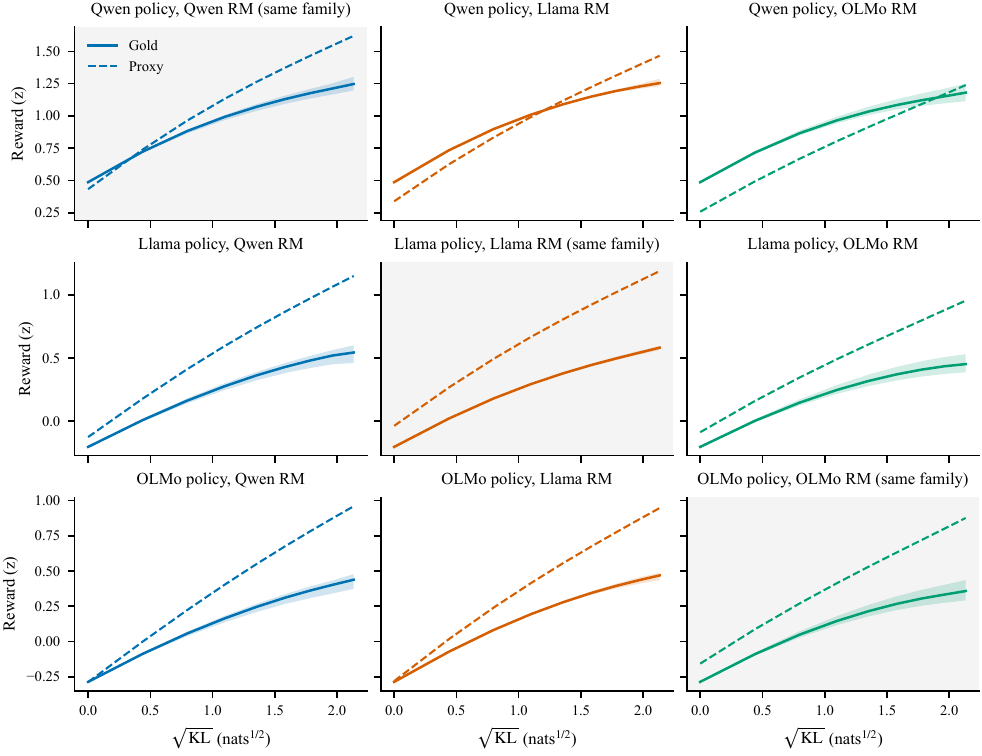}{0.9\textwidth}
\caption{BoN in the $3\times3$ grid, matched tier. Proxy and gold reward against $\sqrt{\mathrm{KL}}$ for same-family and cross-family cells.}
\label{fig:bon}
\end{figure*}

\section{Audit Calibration and Analysis Plan}\label{app:audit}

\begin{figure*}[!ht]
\centering
\fig{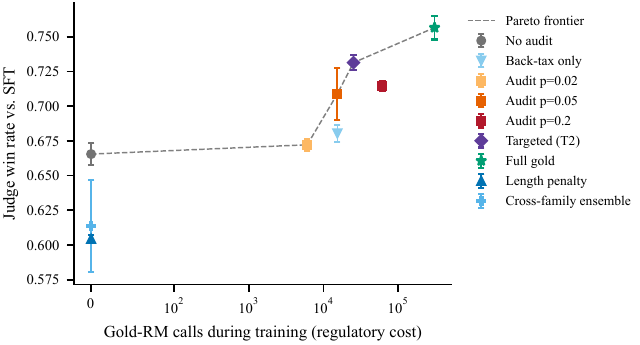}{0.72\textwidth}
\caption{Judge win rate against the number of gold-reward calls during training. The dashed line is the cost frontier.}
\label{fig:pareto}
\end{figure*}

Table~\ref{tab:audit} lists every audit condition and baseline at step 600, and Figure~\ref{fig:pareto} plots the judge win rate against the number of gold-reward calls.

\begin{table*}[!ht]
\centering
\small
\begin{tabular}{lcccccc}
\toprule
Condition & Audit rate & Judge win & Gold (z) & Capture idx. & DWL & Length \\
\midrule
No audit (3) & 0.000\,{\scriptsize$\pm$0.000} & 0.666\,{\scriptsize$\pm$0.008} & +1.67\,{\scriptsize$\pm$0.02} & 0.41\,{\scriptsize$\pm$0.01} & 0.03\,{\scriptsize$\pm$0.01} & 424\,{\scriptsize$\pm$15} \\
Back-tax only (3) & 0.050\,{\scriptsize$\pm$0.001} & 0.680\,{\scriptsize$\pm$0.006} & +1.73\,{\scriptsize$\pm$0.02} & 0.37\,{\scriptsize$\pm$0.01} & 0.02\,{\scriptsize$\pm$0.01} & 423\,{\scriptsize$\pm$9} \\
Audit p=0.02 (3) & 0.020\,{\scriptsize$\pm$0.000} & 0.672\,{\scriptsize$\pm$0.004} & +1.72\,{\scriptsize$\pm$0.03} & 0.20\,{\scriptsize$\pm$0.03} & 0.00\,{\scriptsize$\pm$0.00} & 479\,{\scriptsize$\pm$12} \\
Audit p=0.05 (3) & 0.050\,{\scriptsize$\pm$0.001} & 0.709\,{\scriptsize$\pm$0.019} & +1.71\,{\scriptsize$\pm$0.05} & 0.21\,{\scriptsize$\pm$0.02} & 0.01\,{\scriptsize$\pm$0.01} & 416\,{\scriptsize$\pm$26} \\
Audit p=0.2 (3) & 0.200\,{\scriptsize$\pm$0.001} & 0.714\,{\scriptsize$\pm$0.004} & +1.90\,{\scriptsize$\pm$0.03} & 0.11\,{\scriptsize$\pm$0.04} & 0.00\,{\scriptsize$\pm$0.00} & 504\,{\scriptsize$\pm$19} \\
Targeted (T2) (3) & 0.082\,{\scriptsize$\pm$0.002} & 0.731\,{\scriptsize$\pm$0.005} & +1.67\,{\scriptsize$\pm$0.04} & 0.05\,{\scriptsize$\pm$0.03} & 0.00\,{\scriptsize$\pm$0.00} & 470\,{\scriptsize$\pm$38} \\
Full gold (2) & 1.000\,{\scriptsize$\pm$0.000} & 0.757\,{\scriptsize$\pm$0.009} & +2.24\,{\scriptsize$\pm$0.01} & $-0.26$\,{\scriptsize$\pm$0.02} & 0.00\,{\scriptsize$\pm$0.00} & 525\,{\scriptsize$\pm$7} \\
Length penalty (2) & 0.000\,{\scriptsize$\pm$0.000} & 0.604\,{\scriptsize$\pm$0.003} & +1.27\,{\scriptsize$\pm$0.00} & 0.60\,{\scriptsize$\pm$0.02} & 0.05\,{\scriptsize$\pm$0.00} & 152\,{\scriptsize$\pm$0} \\
Cross-family ensemble (2) & 0.000\,{\scriptsize$\pm$0.000} & 0.614\,{\scriptsize$\pm$0.033} & +1.78\,{\scriptsize$\pm$0.02} & 0.22\,{\scriptsize$\pm$0.03} & 0.02\,{\scriptsize$\pm$0.03} & 728\,{\scriptsize$\pm$44} \\
\bottomrule
\end{tabular}
\caption{Audit conditions and baselines at step 600, mean $\pm$ standard deviation over seeds (number of seeds in parentheses). Judge win is the share of prompts on which the averaged judges prefer the final checkpoint over the SFT response. Gold reward and the capture index are partly optimized under audits.}
\label{tab:audit}
\end{table*}

At step 150 the rent was 0.264 without audits and $-0.091$ with full gold reward. Expected fines of 0.1, 0.3 and 1.0 left rents of 0.204, 0.191 and 0.106, which closed 17\%, 20\% and 44\% of the gap. Length penalties of 1, 3, 10 and 1.3 produced mean lengths of 272, 116, 44 and 230 tokens, against 240 for SFT. Realized audit rates matched their targets (0.020, 0.050 and 0.200 on the line, 0.050 for the back tax and 1.0 for full gold), except for T2 at 0.084.

The bunching test uses the disagreement scores of all training samples after step 100, bins of width 0.05 within 1.0 of the threshold, and a polynomial of degree 7 fitted outside a window from 0.25 below the threshold to the threshold. Excess mass is the observed minus fitted count inside the window, divided by the fitted count. A random-audit run in the same cell serves as a placebo, because it has no threshold to avoid. Both judges labeled all 7{,}200 audit pairs and agreed on 87.9\% of them, and the primary outcome averages their probabilities. Each judge alone reproduces the direction and significance of RQ3a and the null result of RQ3b (T2 minus random: 0.030 [$-0.012$, 0.068] and 0.015 [$-0.022$, 0.046]). Claude Haiku 5.5 judged only 23 prompts per run, and its intervals are about six times wider. The bunching test does not depend on any judge, and we compute its intervals from a separate random stream with 10{,}000 draws.

\end{document}